\documentclass[10pt,twocolumn,letterpaper]{article}

\usepackage{iccv}
\usepackage{times}
\usepackage{epsfig}
\usepackage{graphicx}
\usepackage{amsmath}
\usepackage{amssymb}

\usepackage{booktabs}
\usepackage{multirow}
\usepackage{caption}
\usepackage{subcaption}
\usepackage[pagebackref,breaklinks,colorlinks]{hyperref}
\usepackage[capitalize]{cleveref}
\crefname{section}{Sec.}{Secs.}
\Crefname{section}{Section}{Sections}
\Crefname{table}{Table}{Tables}
\crefname{table}{Tab.}{Tabs.}

\iccvfinalcopy 

\def\iccvPaperID{7321} 

\ificcvfinal\fi

\begin{document}

\title{Hypergraph Transformer for Skeleton-based Action Recognition}


\author{Yuxuan Zhou\textsuperscript{$1$},
Zhi-Qi Cheng\textsuperscript{$2$},
Chao Li\textsuperscript{$3$},
Yanwen Fang\textsuperscript{$4$},
Yifeng Geng\textsuperscript{$3$},
Xuansong Xie\textsuperscript{$3$},
Margret Keuper\textsuperscript{$5$}\\
\textsuperscript{1}{University of Mannheim}
\textsuperscript{2}{Carnegie Mellon University}\\
\textsuperscript{3}{Alibaba Group}
\textsuperscript{4}{The University of Hong Kong}
\textsuperscript{5}{University of Siegen}\\
{\tt \small zhouyuxuanyx@gmail.com, zhiqic@cs.cmu.edu,
lllcho.lc@alibaba-inc.com, u3545683@connect.hku.hk}\\
{\tt \small cangyu.gyf@alibaba-inc.com, xingtong.xxs@taobao.com,
keuper@mpi-inf.mpg.de
}
}

\maketitle
\ificcvfinal\thispagestyle{empty}\fi

\begin{abstract}
Skeleton-based action recognition aims to recognize human actions given human joint coordinates with skeletal interconnections.
By defining a graph with joints as vertices and their natural connections as edges, previous works successfully adopted Graph Convolutional networks (GCNs) to model joint co-occurrences and achieved superior performance. More recently, a limitation of GCNs is identified, i.e., the topology is fixed after training. To relax such a restriction, Self-Attention (SA) mechanism has been adopted to make the topology of GCNs adaptive to the input, resulting in the state-of-the-art hybrid models. Concurrently, attempts with plain Transformers have also been made, but they still lag behind state-of-the-art GCN-based methods due to the lack of structural prior. Unlike hybrid models, we propose a more elegant solution to incorporate the bone connectivity into Transformer via a graph distance embedding. Our embedding retains the information of skeletal structure during training, whereas GCNs merely use it for initialization. 
More importantly, we reveal an underlying issue of graph models in general, i.e., pairwise aggregation essentially ignores the high-order kinematic dependencies between body joints. 
To fill this gap, we propose a new self-attention (SA) mechanism on hypergraph, termed Hypergraph Self-Attention (HyperSA), to incorporate intrinsic higher-order relations into the model. 
We name the resulting model \textbf{Hyperformer}, and it beats state-of-the-art graph models w.r.t.~accuracy and efficiency on NTU RGB+D, NTU RGB+D 120, and Northwestern-UCLA datasets. Our code is available at \href{https://github.com/ZhouYuxuanYX/Hypergraph-Transformer-for-Skeleton-based-Action-Recognition}{Hyperformer Github Repository}.
\end{abstract}


\begin{figure}[]
  \centering
      \centering
      \includegraphics[width=0.48\textwidth]{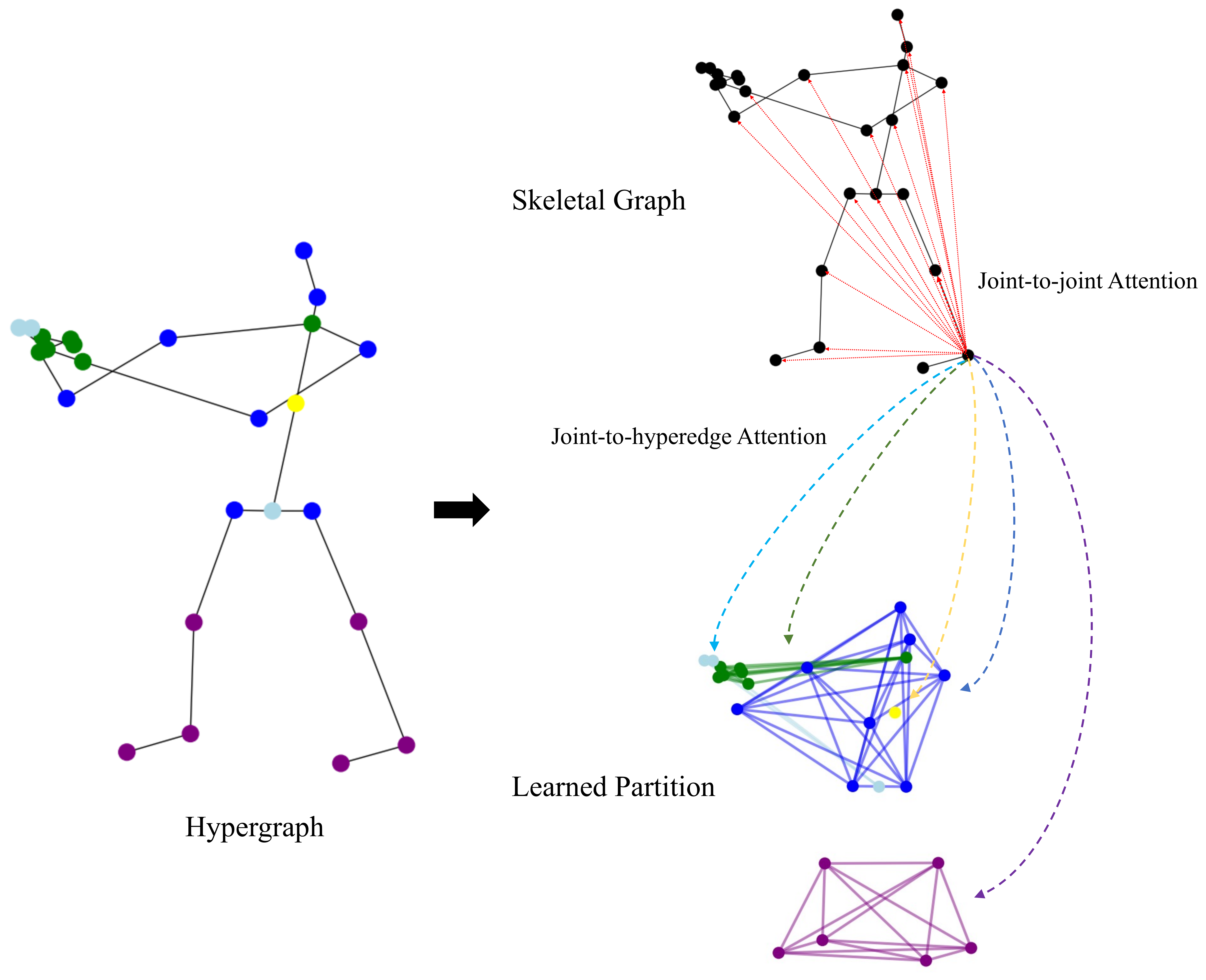}
     \caption{Illustration of our HyperSA using a frame from the action class ``Clapping Hands".
     HyperSA accommodates the additional high-order relations besides the skeletal interconnections.}
     \label{fig:tease}
\end{figure}

\section{Introduction}
\label{sec:intro}
Skeleton-based human action recognition has attracted increasing attention due to its computational efficiency and robustness to environmental variations and camera viewpoints.
One of the key advantages of skeleton-based action recognition is that body keypoints can be easily acquired using sensors such as Kinect \cite{zhang2012microsoft}  or reliable pose estimation algorithms \cite{cao2017realtime}. 
This offers a more reliable alternative to traditional RGB or depth-based methods, making it a promising solution for various real-world applications.

Graph Convolution Networks (GCNs) have been widely used for modeling off-grid data.
To our knowledge, Yan et al.~\cite{yan2018spatial} were the first to treat joints and their natural connections as nodes and edges of a graph, and employ a GCN \cite{kipf2016semi} on such a predefined graph to learn joint interactions.
Since then, GCNs have become the de facto standard of choice for skeleton-based action recognition.
To further capture the interactions between physically unconnected joints, state-of-the-art GCNs \cite{chi2022infogcn, chen2021channel, ye2020dynamic, shi2019two, song2021constructing, song2020stronger} adopt a learnable topology which only uses the physical connections for initialization.
Even so, they still need to rely on attention mechanisms to relax the restriction of the fixed topology, which is the key to the improved performances, as shown in their ablation studies.



Given these facts, it is natural to question whether a purely attention-based Transformer model would be a better candidate for skeleton-based action recognition. However, current research \cite{plizzari2021spatial, shi2020decoupled} has shown that the performance of such models is far from satisfactory. This can be attributed to the fact that the formulation of the vanilla Transformer ignores the unique characteristics of skeleton data, i.e., 
the permutation invariant attention operation is agnostic to the bone connectivity between human body joints. To address this issue, absolute positional embeddings have been used~\cite{vaswani2017attention, dosovitskiy2021an, touvron2021training} , but they still lack the necessary structural information. In contrast, relative positional embeddings have been shown to be more effective for Transformers in various tasks, involving language~\cite{shaw2018self, dai2019transformer, he2020deberta}, vision \cite{liu2021swin, zhou2022sp, wu2021rethinking, zhou2022sp}, and graph data \cite{ying2021transformers}. To incorporate the information of bone connectivity of human body as well, we  also introduce a powerful relative positional embedding based on graph distance, inspired by Spatial Encoding in \cite{ying2021transformers}. Our embedding retains the information of skeletal structure during the entire training process, whereas GCNs merely use it for initialization. 

More importantly, we reveal an underlying issue of graph models for this task in general.
For human actions, each type of body joint has its own unique physical functionality. As a result, certain re-occurring groups of body joints are often involved in specific actions, such as the subconscious hand movement for maintaining balance. 
Vanilla attention is not capable of capturing these underlying relationships that are independent from joint coordinates and go beyond pair-wise interactions. 
To compensate for this, we employ the concept of hypergraph \cite{zhou2006learning, feng2019hypergraph, bai2021hypergraph} to accommodate the higher-order relations of body joints. With the hypergraph representation, we propose a novel variant of Self-Attention (SA) called \textbf{Hypergraph Self-Attention (HyperSA)}, which considers both pair-wise and high-order relations. As shown in \cref{fig:tease}, given a partition of the human body joints into different groups, a representation of each group is derived based on its assigned joints. The group representation is then linearly transformed and multiplied with joint queries, allowing joint-to-group interactions in addition to the vanilla joint-to-joint SA. 
Though HyperSA works well with empirical partitions, we additionally propose an approach to search the optimal partition strategy automatically, further improving its performance.

At the same time, Transformers spend a large portion of their capacity on intra-token modeling in their feed-forward layer. While this is important for complex tokens such as image patches or word embeddings, we analyze that such an expensive step is not necessary for joint coordinates which are merely three-dimensional. This implies that the modeling of inter-token relations, or the so-called joint co-occurrences, is the key to successful action recognition. We thus suggest removing MLP layers for computation and memory reduction, and show in \cref{sec:design} that MLP layers are indeed negligible.
This leads to a light-weighted Transformer which is comparable to GCNs in model size and computation cost.



Our main contributions can be summarized as follows:
\begin{itemize}
    \item We propose to incorporate the structural information of human skeleton into Transformer via a relative positional embedding based on graph distance, leveraging the gap between Transformer and state-of-the-art hybrid models. 
    
    \item We devise a novel variant of Self-Attention (SA) called \textbf{Hypergraph Self-Attention (HyperSA)}, based on the hypergraph representation. 
    To our best knowledge, our work is the first to apply the hypergraph representation for skeleton-based action recognition, which considers the pair-wise as well as high-order joint relations, going beyond current state-of-the-art methods. 
    
    \item We construct a light-weighted Transformer based on our proposed relative positional embedding and HyperSA. It beats state-of-the-art graph models on skeleton-based action recognition benchmarks w.r.t.~both efficiency and accuracy.
\end{itemize}

\section{Related work}
In this section, we highlight the most related work to ours regarding skeleton representation as well as the spectacle of application.

\subsection{Representation of skeleton data}
\noindent \textbf{Graph Representation} Graph is the most prevalent choice for representing non-euclidean data and human skeleton can be naturally represented as graph. 
Comparing to other graph models \cite{xu2018how, gilmer2017neural, velivckovic2017graph}, the Graph Convolutional Network (GCN) proposed by Kipf \etal~\cite{kipf2016semi} is widely adopted for action recognition due to its simplicity and thus higher resistance to overfitting.
Recently, Graphformer \cite{ying2021transformers} achieves state-of-the-art performance on generic graph data. It reveals the necessity of effectively encoding
the structural information into Transformers. Their finding conforms to our analysis w.r.t.~the aspect of topology, and we analogously propose a powerful k-Hop RPE for our Hyperformer.   

\noindent \textbf{Hypergraph Representation} In real-world scenarios,
relationships could go beyond pairwise associations. Hypergraph further considers higher-order correlations among data. Although hypergraphs can be modeled as
a graph approximately via techniques such as clique expansion\cite{zhou2006learning}, such approximations fail to capture higher-order relationships in the data and result
in unreliable performance~\cite{chien2019hs, li2017inhomogeneous}.
This motivates
the study of learning on hypergraphs \cite{hein2013total, zhang2017re, feng2019hypergraph, yadati2019hypergcn}. Attention-based hypergraph models have also been proposed for multi-modal learning \cite{kim2020hypergraph} and inductive text classification \cite{ding2020more}. Our proposed HyperSA is the first hypergraph attention which is specially designed for skeleton-based action recognition.

\subsection{Skeleton-based action recognition}
In early years, RNNs \cite{du2015hierarchical, song2017end,zhang2017view}
have been a popular choice to tackle the problem of skeleton-based human action recognition.
The application of CNNs for this task 
\cite{ke2017a, liu2017enhanced}
is also well studied.  
Nevertheless, the spatial interactions between joints are ignored in the above methods, 
and GCNs have become a more common choice in this field, by successfully modelling the spatial configurations
as graphs. 

\textbf{GCN-based approaches}  Yan \etal~\cite{yan2018spatial} first introduced GCN \cite{kipf2016semi} to model the joint correlations and demonstrated its effectiveness for action recognition. However, 
the limitation of assuming a fixed topology according to the natural connections is identified later, and most follow-up works adopt a learnable topology for action recognition.
Many among them \cite{cheng2020decoupling, shi2019two, chen2021channel, ye2020dynamic} also employ attention or similar mechanisms to produce a data-dependent component of the  
topology (analogous to Graph Attention Networks \cite{velivckovic2017graph}), boosting GCN's performance further.

\textbf{Transformer-based approaches}
Attempts to tackle this problem with Transformers have been made recently.
They mainly focus on handling the challenge brought by the extra temporal dimension. 
\cite{plizzari2021spatial} propose a two-stream model consisting of spatial and temporal Self-Attention for modeling intra- and inter-frame correlations, respectively. Instead, \cite{shi2020decoupled} employ a Transformer which models the spatial and temporal dimension in an alternate fashion. 
Nevertheless, none of them achieved comparable results to state-of-the-art GCN-based approaches. Since they simply follow the design of vanilla Transformers, the special characteristics of skeleton data are ignored.

\section{Preliminaries}
In this section, we recap the definition of Self-Attention and hypergraphs.

\subsection{Self-Attention}



Given an input sequence in the form of $X=(\vec{x}_1,...,\vec{x}_n)$, each token $\vec{x}_i$ is first projected into \textit{Key} $\vec{k}_i$ , \textit{Query} $\vec{q}_i$  and \textit{Value} $\vec{v}_i$  triplets.
Then the so-called attention score $A_{ij}$ between two tokens is obtained by applying a softmax function to the dot product of $\vec{q}_i$ and $\vec{k}_j$ \cite{vaswani2017attention}:
\begin{equation}\label{eq:1} 
 A_{ij} = \vec{q}_i \cdot \vec{k}_j^\top,
\end{equation}
the final output at each position is computed as the weighted sum of all Values:
\begin{equation}
\label{eq:2}
\vec{y}_i = \sum_{j=1}^n A_{ij}\vec{v}_j
.
\end{equation}
An extension called Multi-Head Self-Attention (MHSA) is often adopted by Transformers in practice. It
divides the channel dimension into subgroups and apply Self-Attention to each subgroup in parallel to learn different kinds of inter-dependencies. For simplicity, we omit the notation of MHSA in this paper.
\subsection{Hypergraph representation}

Unlike standard graph edges, a hyperedge in a hypergraph connects
two \emph{or more} vertices. An unweighted hypergraph is defined as $\mathcal{H} =
(\mathcal{V}, \mathcal{E})$, which consists of a vertex set $\mathcal{V}$ and a hyperedge set $\mathcal{E}$. The hypergraph $\mathcal{H}$ can be denoted by a $\vert \mathcal{V} \vert \times \vert \mathcal{E} \vert $ incidence matrix $H$, with entries defined as follows:
\begin{equation}
\label{eq:3}
h_{v, e}=\left\{
	\begin{aligned}
	1, \quad if \quad v \in e\\
	0, \quad if \quad v \notin e\\
	\end{aligned}\right
	.
\end{equation}
The degree of a node $ v \in \mathcal{V}$ is defined as $ d(v) =
\sum_{e \in \mathcal{E}} h_{v, e}$, and the degree of a hyperedge $e \in \mathcal{E}$ is defined
as $d(e) = \sum_{v \in \mathcal{V}} h_{v, e}$. The degree matrices $D_e$ and $D_v$ are constructed by setting all the edge degrees and all the vertex degrees as their diagonal entries,
respectively.

In this work, we consider the special case of $d(v)=1$ for all vertices, i.e.~, body joints are divided into $\vert \mathcal{E} \vert$ disjoint subsets, which is efficient in practice. Notably, the incidence matrix $H$ is equivalent to a partition matrix in this case.
Each row is a one hot vector denoting the group to which each joint belongs.


\begin{figure*}[ht]
  \centering

      \includegraphics[width=0.9\textwidth]{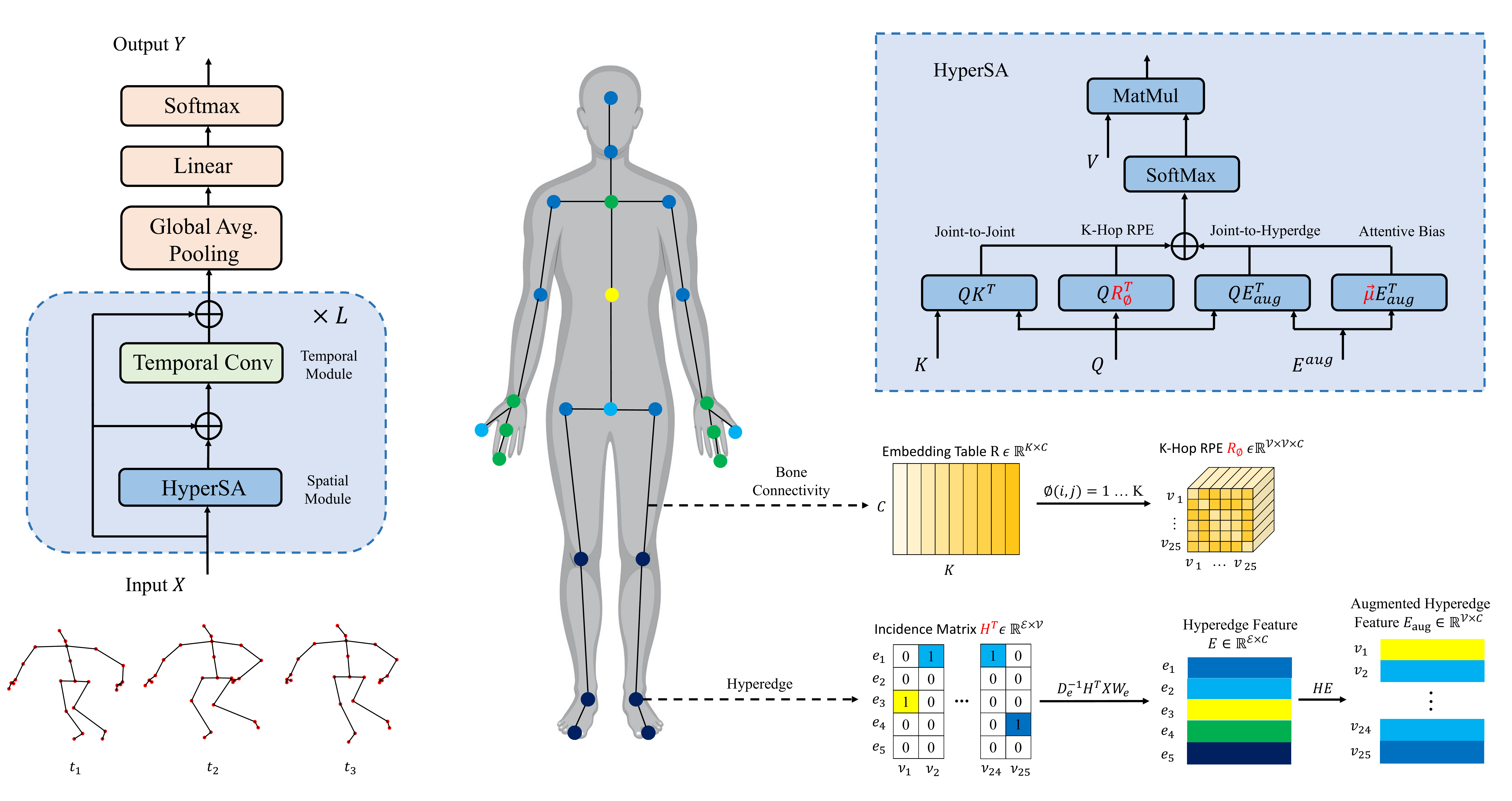}
     \caption{Model architecture overview and illustration of our proposed HyperSA layer.}
     \label{fig:model}
\end{figure*}

\section{Method}
As analyzed in \cref{sec:intro}, multiple specific joints often move cooperatively in an action, \ie~there are inherent higher-order relations beyond the pair-wise relation. Therefore, we propose to introduce the prior information of the intrinsic hyper connections into vanilla Self-Attention. Specifically, a novel \textbf{Hypergraph Self-Attention (HyperSA)} layer is introduced, which makes Transformers aware of extra higher-order relations shared by a subset of joints connected to each hyperedge. 

\subsection{Deriving the hyperedge feature}

Given an incidence matrix $H$, we propose an effective approach to obtain the feature representation for each subset of joints connected to a hyperedge. 
Let $C$ denote the number of feature dimensions, individual joint features $X \in \mathbb{R}^{\vert \mathcal{V} \vert \times C}$ are first aggregated into subset representations $E \in \mathbb{R}^{ \vert \mathcal{E} \vert \times C}$ by the following rule:
\begin{equation}
\label{eq:4}
     E = D_e^{-1}H^\top XW_{e},
\end{equation}
where:
\begin{itemize}
    \item The product of incidence matrix $H$ and input $X$ essentially sums up the belonging joint features of each subset. 
    \item The inverse degree matrix of hyperedges are multiplied for the purpose of normalization. 
    \item The projection matrix $W_{e} \in \mathbb{R}^{C\times C}$ further transforms the features of each hyperedge to obtain their final representations.
\end{itemize}
 

Then we construct an augmented hyperedge representation $E_{aug} \in \mathbb{R}^{ \vert \mathcal{V} \vert \times C}$ by assigning hyperedge representations to the position of each associated joint:

\begin{equation}
\label{eq:5}
    E_{aug} = HD_e^{-1}H^\top XW_{e}.
\end{equation}

\subsection{Encoding human skeleton structure}
Human body joints are naturally connected with bones and form, together with the latter, a bio-mechanical model. In such a mechanical system, the movement of each joint in an action is strongly influenced by their connectivities. Therefore, it is beneficial to take the structural information of human skeleton into account.

Analogous to the established Relative Positional Embedding (RPE) for image \cite{wu2021rethinking} and language \cite{he2020deberta, shaw2018self} Transformer, we propose a powerful k-Hop Relative Positional Embedding $R_{ij} \in \mathbb{R}^C$, which is indexed from a learnable parameter table by the Shortest Path Distance (SPD) between the $i^{th}$ and $j^{th}$ joints. 
In comparison to the learnable scalar spatial encoding in \cite{ying2021transformers}, it has larger capacity and interacts with the query additionally.

\subsection{Hypergraph Self-Attention}
With the obtained hyperedge representation and skeleton topology encoding, we now define our Hypergraph Self-Attention as follows:
\begin{align}
\label{eq:6}
\begin{split}
     A_{ij} = & \underbrace{\vec{q}_i \cdot \vec{k}_j^\top}_{\text{(a)}} + \underbrace{\vec{q}_i \cdot E_{aug, j}^\top}_{\text{(b)}} \\
     & + \underbrace{\vec{q}_i \cdot R_{\phi(i, j)}^\top}_{(c)} + \underbrace{\vec{u} \cdot E_{aug, j}^\top}_{(d)},
    \end{split}
\end{align}
where $\vec{u} \in \mathbb{R}^{C} $ is a learnable static key regardless of the query position. 
\begin{itemize}
    \item Term (a) alone is the vanilla SA, which represents joint-to-joint attention.
    \item Term (b) computes the joint-to-hyperedge attention between the $i^{th}$ query and the corresponding hyperedge of the $j^{th}$ key.
    \item Term (c) is the term for injecting the structural information of human skeleton with k-Hop Relative Positional Embedding. 
    \item Term (d) is intended for calculating the attentive bias of different hyperedges independent of the query position. It assigns the same amount of attention to each joint connected to a certain hyperedge.
\end{itemize}

Note that terms (a) and (b) can be combined by distributive law and require merely an extra step of matrix addition. Moreover, term (d) has $O(\vert \mathcal{V} \vert C^2)$ complexity and thus requires negligible computation in comparison to term~(a).

\noindent \textbf{Relational Bias}
Transformers assume the input tokens to be homogeneous, whereas human body joints are inherently heterogeneous, \eg each physical joint plays a unique role and thus has different relations to others.
Taking the heterogeneity of the skeleton data into account, we propose to represent the inherent relation of each joint pair as a scalar trainable parameter $B_{ij}$, called Relational Bias (RB).  It is added to the attention scores before aggregating the global information:
\begin{equation}
\label{eq:7}
\vec{y}_i = \sum_{j=1}^n (A_{ij} + B_{ij}) \vec{v}_j
,
\end{equation}

\subsection{Partition strategy}
\label{sec:learn}

\begin{figure}[]
  \centering
  \begin{subfigure}[t]{0.18\textwidth}
      \centering
      \includegraphics[width=0.75\textwidth]{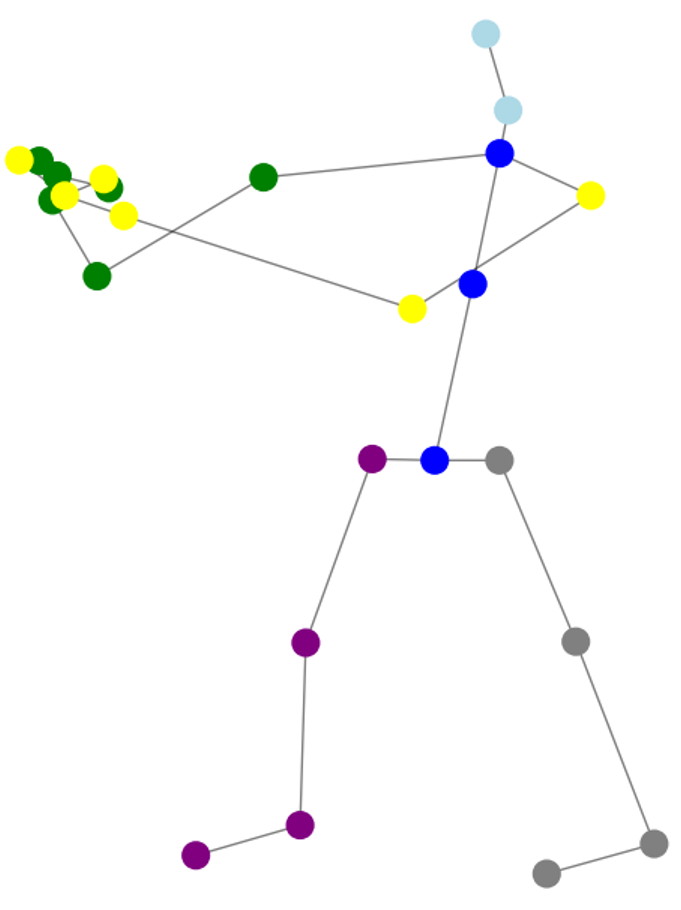}
      \caption{Empirical partition a.}
      \label{fig: 2}
  \end{subfigure}
    \hspace{1.5cm}
  \begin{subfigure}[t]{0.18\textwidth}
      \centering
      \includegraphics[width=0.75\textwidth]{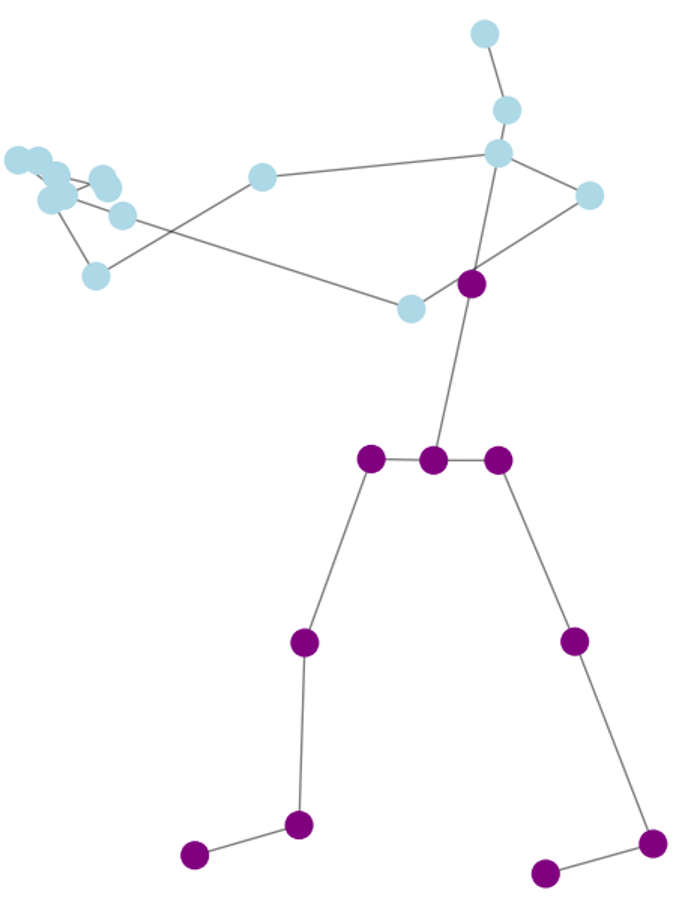}
      \caption{Empirical partition b.}
      \label{fig:3}
  \end{subfigure}
  \hspace{0.6cm}
  \begin{subfigure}[t]{0.18\textwidth}
      \centering
  \includegraphics[width=0.75\textwidth]{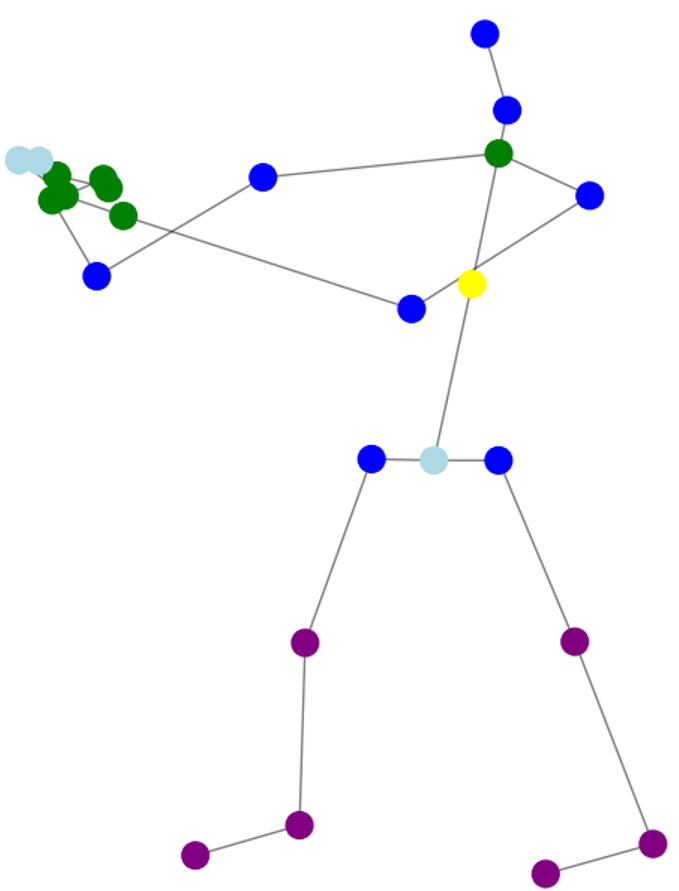}
      \caption{Learned partition a.}
      \label{fig:4}
  \end{subfigure}
        \hspace{1.5cm}
  \begin{subfigure}[t]{0.18\textwidth}
      \centering
  \includegraphics[width=0.75\textwidth]{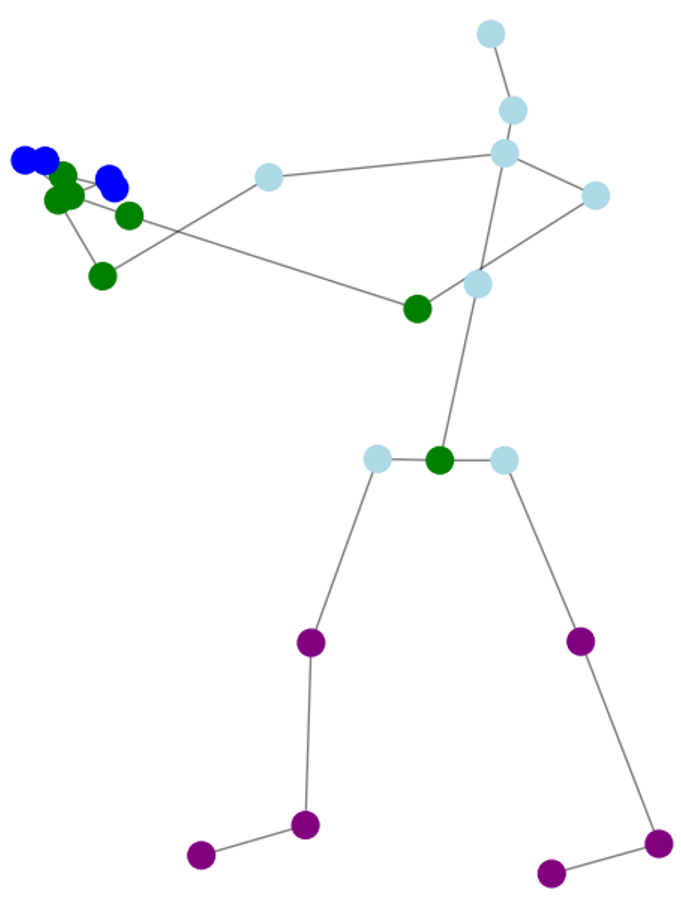}
      \caption{Learned  partition b.}
      \label{fig:leg}
  \end{subfigure}
     \caption{Visualization of the empirical and learned partitions. Different node colors stand for different subgroups for each partition strategy.}
     \label{fig:2}
\end{figure}

Empirically, human skeletons could be divided into a number of body parts, which have been well studied in previous work~\cite{thakkar2018part, huang2020part, song2020stronger}.
We experimentally show that our Hyperformer with empirical partitions yields excellent performance. 
However, finding an optimal empirical partition strategy is laborious and the optimal partition strategy is restricted to a certain skeleton with a fixed number of recorded joints. In this work, we also provide an approach to automate the search process for an effective partition strategy.

To make the partition matrix learnable, we parameterize and relax the binary partition matrix to its continuous version by applying a softmax along its column axis:
\begin{equation}
    \Tilde{H} = \{ \Tilde{h}_{ve} = \frac{{\rm exp} (h_{ve})} {\sum^{\vert \mathcal{E}\vert}_{e=1} {\rm exp} (h_{ve})}; i=1...\vert \mathcal{V} \vert, j=1...\vert \mathcal{E} \vert \}.
\end{equation}
The problem of finding an optimal discrete partition matrix $H$ is thus reduced to learning an optimal continuous partition matrix $\Tilde{H}$, which can be optimized jointly with Transformer parameters. 




At the end of the optimization, a discrete partition matrix can be obtained by applying an argmax operation along each row of $\Tilde{H}$:
\begin{equation}
    H = argmax(\Tilde{H}).
\end{equation}
Note that a number of different proposals can be easily acquired by varying the initialization of $\Tilde{H}$. We experimentally show that all the proposals prove to be reasonable. Interestingly, all the learned proposals are symmetric as shown in \cref{fig:2}, indicating that symmetry is an important aspect of inherent joint relations. 

\begin{figure*}[ht]
  \centering
      \includegraphics[width=0.65\textwidth]{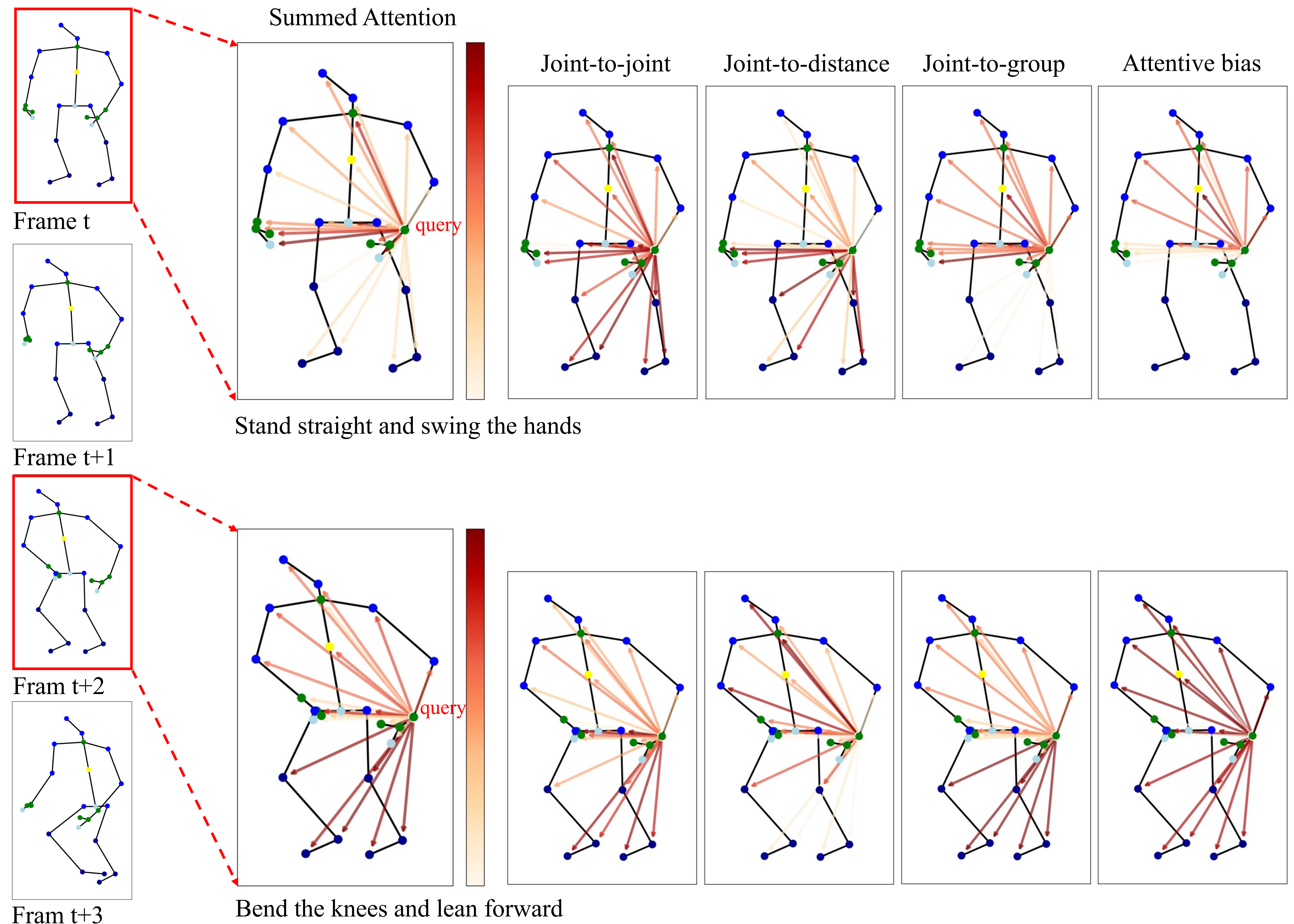}
     \caption{Visualization of the attention scores for the action class "Jump Up". The directed edges represent the attention weights w.r.t.~the query joint of left wrist and range from light orange to dark red with the increase of the weights. The black edges stand for the bones and the joints are assigned different colors according to their connected hyperedges as in \cref{fig:2} (c). }
     \label{fig:attention}
\end{figure*}

\subsection{Model architecture}

We first revisit the architectural design of Transformers for skeleton data.
Then we built our Hyperformer based on our analysis. HyperSA is employed for spatial modeling of each frame and a lightweight convolutional module is adopted for temporal modeling, following the design of state-of-the-art models \cite{chi2022infogcn, chen2021channel} in this field. 


\noindent \textbf{Spatial Modeling}
We apply Layer Normalization (LN) before the multi-head HyperSA and add a residual connection to the output, following the standard Transformer architecture \cite{vaswani2017attention}.
Based on our analysis in \cref{sec:intro}, we further remove the Multi-Layer-Perceptron (MLP) layers. To introduce non-linearity, a ReLU layer is added after each block of spatial and temporal modeling modules instead.


\noindent \textbf{Temporal Modeling}
To model the temporal correlation of
the human pose, we adopt the Multi-Scale Temporal Convolution (MS-TC) module \cite{chen2021channel, liu2020disentangling, chi2022infogcn} for our final model. This module contains three convolution branches with a 1 × 1 convolution to reduce channel dimension, followed by different combinations of kernel sizes
and dilations. The outputs of convolution branches are
concatenated.

Hyperformer is constructed by stacking HyperSA and Temporal Convolution layers alternately as follows:
\begin{align}
    & z^{(l)} = \text{HyperSA}(LN(z^{(l-1)})) + z^{(l-1)} \\
    & z^{(l)} = \text{TemporalConv}(LN(z^{(l)})) + z^{(l-1)} \\
    & z^{(l)} = \text{ReLU}(z^{(l)})
\end{align}







\section{Experiments}
In this section, we first compare our Hyperformer to  state-of-the-art approaches on skeleton-based human action recognition benchmarks and show the superior performance of our model. 
Then we conduct an ablation study for a deeper understanding of our proposed HyperSA. Finally, we evaluate our approach qualitatively by visualizing each component of HyperSA.

\subsection{Experimental settings}
\subsubsection{Datasets}
\label{data}
We evaluate our proposed Hyperformer on three commonly adopted public datasets NTU-RGB+D \cite{shahroudy2016ntu}, NTU-RGB+D120 \cite{liu2019ntu} and Northwestern-UCLA \cite{wang2014cross} which are briefly introduced in the following. 


\noindent
\textbf{NTU RGB+D} 
\cite{shahroudy2016ntu} is a widely used public dataset for skeleton-based human action recognition. It contains
56880 skeletal action sequences. 
Each action sequence is conducted by one or two subjects and captured by three Microsoft Kinect-V2 depth sensors set at the same height but from different horizontal viewpoints simultaneously. 
There are two benchmarks for evaluation including Cross-Subject (X-Sub)
and Cross-View (X-View) settings. For the X-Sub, the training and test sets come from two disjoint sets of 20 subjects each. 
For X-View, the training set contains 37920 samples captured by the camera views 2 and 3, and the test set includes 18960 sequences captured by camera view 1. 

\noindent
\textbf{NTU RGB+D 120} \cite{liu2019ntu} is an extension of NTURGB+D dataset with 57367 additional skeleton sequences over 60 additional action classes.
It is currently the largest available dataset with 3D joint annotations for human action recognition and contains 32 setups, each of which represents a different location and background. Two benchmark evaluations were suggested by the author including Cross-Subject (X-Sub) and Cross-Setup (X-Setup) settings. 


\noindent
\textbf{Northwestern-UCLA} \cite{wang2014cross} dataset is
recorded by three Kinect sensors from different viewpoints. It includes 1494 video sequences of 10 action categories. 


\subsubsection{Implementation details}
\label{implement}
All experiments are conducted with the PyTorch \cite{paszke2019pytorch} deep learning library.
We train the model for a total number of 140 epochs with standard cross-entropy loss.
The learning rate is initialized to 0.025 and reduced at 110 and 120 epochs by 0.1, basically following the strategy in \cite{chi2022infogcn}.
For NTU RGB+D and NTU RGB+D 120, the batch size is set to 64, each sample is resized to 64 frames, and we adopt the code of \cite{zhang2020semantics} for data pre-processing.
For Northwestern-UCLA, we use a batch size of 16, and
follow the data pre-processing in \cite{cheng2020skeleton,chen2021channel}.
Our code is based on the official implementations of \cite{touvron2021training}, \cite{chen2021channel} and \cite{zhang2020semantics}. We employ a model with a total number of 10 layers and 216 hidden channel dimensions for all the experiments.

\begin{table*}[ht]
  \centering
    \caption{Action classiﬁcation performance on the NTU RGB+D and NTU RGB+D 120 dataset. Following the common setup, we report results using 4 modalities for a fair comparison. InfoGCN \cite{chi2022infogcn} reported results using extra MMD losses and we mark them with *.}
    \resizebox{0.86\textwidth}{!}{
  \begin{tabular}{@{}l|lccccc@{}}
    \toprule
   \multirow{2}*{Type} & \multirow{2}*{Methods} & \multirow{2}*{Parameters(M)} &\multicolumn{2}{c}{NTU RGB+D 60} &  \multicolumn{2}{c}{NTU RGB+D 120} \\
         &   &  & X-Sub(\%) & X-View(\%)    &  X-Sub(\%)  & X-Set(\%)   \\
    \midrule
     \multirow{2}*{RNN} & VA-LSTM \cite{zhang2017view} & - & 79.4 & 87.6 & - & - \\
    &  AGC-LSTM \cite{si2019attention} & 22.9M & \underline{89.2} &  \underline{95.0} & - & - \\ 
    \hline
       \multirow{2}*{CNN} & VA-CNN (aug.) \cite{zhang2019view} & 24.09M & 88.7 & 94.3 & - & - \\
      & Ta-CNN+ \cite{xu2021topology} & 1.06M & \underline{90.7} & \underline{95.1} & \underline{85.7} & \underline{87.3} \\
      \hline
    \multirow{4}*{GCN}   
    
      & Shift-GCN \cite{cheng2020skeleton} & 2.8M & 90.7 & 96.5 & 85.9 & 87.6\\
      & DC-GCN+ADG \cite{cheng2020decoupling} & 4.9M & 90.8 & 96.6 & 86.5 & 88.1 \\ 
       & MS-G3D \cite{liu2020disentangling}  & 2.8M & 91.5 & 96.2  & 86.9 & 88.4  \\
      
       & MST-GCN  \cite{chen2021multi} & 12.0M& \underline{91.5} & \underline{96.6} & \underline{87.5} & \underline{88.8}\\
    
       \hline
        \multirow{4}*{\shortstack{Hybrid Model \\ (
        GCN + Att})} 
    
       & Dynamic GCN \cite{ye2020dynamic}  & 14.4M &  91.5 & 96.0 & 87.3 & 88.6 \\
       & EfficientGCN-B4 \cite{song2021constructing} & 2.0M & 91.7 & 95.7 & 88.3 & 89.1  \\
            & InfoGCN (Joint Only) \cite{chi2022infogcn} & 1.6M & 89.8* & 95.2* & 85.1*(84.3) & 86.3* \\
       & InfoGCN \cite{chi2022infogcn} & 1.6M & \underline{92.7*} & \underline{\textbf{96.9*}} &\underline{89.4*}(89.1) & \underline{90.7*} \\
   \hline
   \multirow{4}*{Transformer} 
    & ST-TR \cite{plizzari2021spatial} & 12.1M & 89.9 & 96.1    & 82.7 & 84.7 \\
   & DSTA \cite{shi2020decoupled} & 4.1M & 91.5 & 96.4 & 86.6 & 89.0\\
& Hyperformer (Joint Only) & 2.6M 
& 90.7 & 95.1 & 86.6 & 88.0
\\

     &   \textbf{Hyperformer} & 2.6M & \underline{\textbf{92.9}}  & \underline{96.5} & \underline{\textbf{89.9}} & 
     \underline{\textbf{91.3}} \\
    \bottomrule
  \end{tabular}
  }
  \label{tab:ntu}
\end{table*}

\begin{table}[h]
  \centering
    \caption{Action classiﬁcation performance on the Northwestern-UCLA dataset.}
    \resizebox{0.45\textwidth}{!}{
  \begin{tabular}{@{}l|lcc@{}}
    \toprule
       \multirow{1}*{Type} & \multirow{1}*{Methods} 
          &Acc (\%)  \\
    \midrule
   \multirow{2}*{RNN} &
        TS-LSTM \cite{lee2017ensemble} & 89.2 \\
       &      2s-AGC-LSTM \cite{si2019attention} & \underline{93.3} \\
        \hline 
                  \multirow{2}*{CNN} 
        & VA-CNN (aug.) \cite{zhang2019view} & 90.7 \\
         & Ta-CNN \cite{xu2021topology} & \underline{96.1} \\
         \hline
          \multirow{3}*{GCN} &
      4s-shift-GCN \cite{cheng2020skeleton} & 94.6 \\
      & DC-GCN+ADG \cite{cheng2020decoupling} & 95.3 \\ 
        & InfoGCN \cite{chi2022infogcn} (*with MMD losses)
        & \underline{96.6*} \\
        \hline
           \multirow{1}*{Transformer} & \textbf{Hyperformer} & 
    



           \underline{\textbf{96.7}} \\
    \bottomrule
  \end{tabular}
  }
  \label{tab:ucla}
\end{table}
\subsection{Comparison with state-of-the-art approaches}
Following most recent state-of-the-art approaches \cite{cheng2020skeleton, ye2020dynamic, chen2021channel, chen2021multi}, we adopt a multi-stream fusion strategy, \ie,
there are 4 streams which take different modalities including \textbf{joint}, \textbf{bone}, \textbf{joint motion} and \textbf{bone motion} as input respectively. Joint modality refers to the original skeleton coordinates; bone modality represents the differential of spatial coordinates; joint motion and bone motion modalities use the differential on
temporal dimension of joint and bone modalities, respectively. The softmax scores
of 4 streams are added to obtain the fused score. 


The comparison on the three datasets is shown in \cref{tab:ntu} and \cref{tab:ucla}, respectively. As shown in \cref{tab:ntu}, our model reaches state-of-the-art results on all benchmarks, except the Cross-View Setup of NTU RGB+D dataset. Our Hyperformer performs slightly worse than InfoGCN \cite{chi2022infogcn}. Nevertheless, the latter relies on two additional MMD loss terms and a number of associated hyperparameters, including loss coefficients, noise ratio and z prior gain. NTU RGB+D 120 is a more challenging dataset, on which the results of previous Transformer-based approaches are not satisfying. On the contrary, our model achieves the best results among all model categories. This conforms to the findings that attention mechanism benefits more from a large amount of data than Convolution \cite{dosovitskiy2021an, touvron2021training}. The Northwestern-UCLA dataset is particularly challenging since it contains much fewer training samples, making it even harder for Transformer-based models to compete. With the prior knowledge of bone connectivity and underlying high-order relations of joints, our Hyperformer still yields state-of-the-art results in such a low-data regime.

\subsection{Ablation study}
In this section, we revisit the role of MLP layers for this task and compare Hyperformer with the vanilla Transformer to show the effectiveness of our proposed model. We also analyze the contribution of each component of our HyperSA. In addition,
we compare our learned partition strategy with empirical ones to show its effectiveness.
For the ablation study, all experiments
are conducted on the
X-sub benchmark of NTU RGB+D using the joint modality as input only.


\subsubsection{The design of Hyperformer}
\label{sec:design}
Before replacing standard SA layers with our HyperSA, we removed the MLP layers based on the results in \cref{tab:construct}. As can be seen, our HyperSA layers contribute most to the final performance, with an significant improvement of $2.8\%$ absolute accuracy. The MS-TC module further improves over vanilla TC by $0.4\%$.
In \cref{tab:construct}, it can be seen that our Hyperformer achieves significantly higher accuracy than the vanilla baseline with fewer parameters thanks to the listed design choices. We provide more detailed results in \cref{tab:more}, validating the effectiveness of each individual HyperSA components.




\begin{table}[h]
\caption{Constructing Hyperformer from the vanilla baseline. Note that the adopted MS-TC module for our final model has fewer parameters than vanilla Temporal Convolution (TC) due to the dimension reduction via 1x1 convolutions, see \cref{fig:model}}
  \centering
  \resizebox{0.45\textwidth}{!}{
  \begin{tabular}{@{}lccc@{}}
    \toprule
   Model  & \multicolumn{1}{c}{Parameters} & FLOPs  & Acc(\%) \\
    \midrule
       \multirow{1}*{SA + MLP + TC} & 7.2M & 25.6G &  88.3 \\
        SA + TC & 3.6M & 14.1G & 87.5 \\
        \multirow{1}*{HyperSA + TC}   & 4.1M &  16.7G & 90.3 \\
          \multirow{1}*{HyperSA + MS-TC}   & 2.6M & 14.8G & 90.7 \\
    \bottomrule
  \end{tabular}
  }
  \label{tab:construct}
\end{table}

\begin{table}[h]
\caption{The effectiveness of the HyperSA components.}
  \centering
  \resizebox{0.4\textwidth}{!}{
  \begin{tabular}{@{}lccc@{}}
    \toprule
   Model  &  Acc(\%) \\
    \midrule
        SA + TC  & 87.5 \\
        \multirow{1}*{SA + TC + Joint-to-hyperedge attn}     & 89.6 \\
         \multirow{1}*{SA + TC + K-Hop RPE}   &    89.5 \\
           \multirow{1}*{SA + TC + Hyperedge Attentive Bias}    &  89.6  \\
               \multirow{1}*{Full HyperSA + TC}   &   90.3 \\
    \bottomrule
  \end{tabular}
  }
  \label{tab:more}
\end{table}



\subsubsection{Effect of different partition strategies}
\begin{table}[h]
  \centering
  \caption{The effect of different partition strategies}
  \resizebox{0.3\textwidth}{!}{
  \begin{tabular}{@{}lcc@{}}
    \toprule
    Partition Strategy & Acc(\%) \\
    \midrule
        \multirow{1}*{Body Parts}  & 90.5  \\
    Upper and Lower Body  &  90.4 \\
        \multirow{1}*{Learned a}  & 90.7 \\
          \multirow{1}*{Learned b}  & 90.7 \\
    \bottomrule
  \end{tabular}
  }
  \label{tab:partition}
    \end{table}
In \cref{tab:partition}, we compared empirical partitions according to body parts (see \cref{fig:2}), and our learned partitions. Note that we obtain different partition proposals when the model is trained with different seeds. All proposals prove to be effective.
Overall, Hyperformer delivers stable performance and achieves the best result with a learned partition using the approach described in \cref{sec:learn}.

\subsection{Qualitative results}
In order to showcase the effectiveness of our approach, we visualize the attention scores of our HyperSA as well as the four decomposed terms in \cref{eq:6} at the first layer in \cref{fig:attention}. More specifically, we draw the attention scores  w.r.t.~the query joint of left wrist for two single frames respectively. The directed edges represent the attention weights and range from light orange to dark red with the increase of attention score. The black edges stand for the bones and the joints are assigned different colors according to their connected hyperedges.  

At Frame $t$, the person stands straight and starts to swing the hands, preparing to jump up. 
The joint-to-joint attention is distracted by a large number of joints, whereas the joint-to-group attention concentrates on the upper body. As the sum of the four terms, the final attention reasonably focuses on the hands and neck.

At Frame $t+2$, the person bends the knees and leans the upper body forward to squeeze the leg muscles. Although the joint-to-joint attention is successfully attached to the feet and waist, the knees and heels are less valued. This incomplete and unstable attention is unavoidable due to the pairwise mechanism. However, our proposed joint-to-group attention solves this issue by exploiting the underlying group relation.  

\section{Discussion}
\noindent\textbf{Broader Impact. }
Skeleton-based action recognition is widely studied due to its computation efficiency and robustness against the change of lighting conditions, compared to video-based action recognition. Moreover, skeleton-data better preserves the privacy since the identities of human subjects are removed. 

\noindent\textbf{Limitations.}
\label{limit}
Our paper focuses on making Transformers fit for skeleton-based action recognition and specially considers the intrinsic nature of skeleton data. Therefore, our proposed Hyperformer may not be advantageous for other tasks, but the idea of taking inherent relations into account suggests the potential to better utilize Transformers. 
\section{Conclusion}

In this work, we successfully incorporate the information of skeletal structure into Transformer by proposing a relative positional embedding based on graph distance. This is a more elegant solution than previous hybrid models. Moreover, we identified a limitation of graph models for the task of skeleton-based action recognition, i.e., high-order joint relations are ignored.
Therefore, we propose a novel HyperSA layer to make Transformer models aware of these inherent relations. 
The resulting model called Hyperformer establishes the state-of-the-art performance.

\clearpage

{\small
\bibliographystyle{ieee_fullname}
\bibliography{egbib}
}

\end{document}


\ificcvfinal\thispagestyle{empty}\fi


\begin{document}
 
\appendix

\section{Python implementation}
We list our Pytorch implementation of HyperSA layer in \cref{toa}. For simplicity, we omit the code for initialization. 


\begin{listing*}[h]%
\caption{{\tt HyperSA.py}}%
\label{toa}%
\begin{lstlisting}[language=Python]
import torch
from torch import nn
class HyperSA(nn.Module):
    def __init__(self, dim_in, dim, num_heads=9, qkv_bias=False, H=None, qk_scale=None, attn_drop=0., proj_drop=0., hops=None, num_point=25, **kwargs):
        '''
        :param H: Incidence Matrix, shape (num_point, num_hyperedge)
        :param hops: Shortest Path Distance (SPD), shape (num_point, num_point) 
        
        '''
        super().__init__()
        self.num_heads = num_heads
        self.dim = dim
        head_dim = dim // num_heads
        self.scale = qk_scale or head_dim ** -0.5
        self.num_point = num_point
        self.rpe_table = nn.Parameter(torch.zeros((hops.max()+1, dim)))
        self.u = nn.Parameter(torch.zeros(num_heads, head_dim))
        self.relational_bias = nn.Parameter(torch.stack([torch.eye(num_point]) for _ in range(num_heads)], dim=0), requires_grad=True)
        self.qkv = nn.Conv2d(dim_in, dim * 3, 1, bias=qkv_bias)
        self.proj = nn.Conv2d(dim, dim, 1)
        self.e_proj = nn.Conv2d(dim_in, dim, 1, bias=False)
        self.H = H


    def forward(self, x, joint_label, groups, pe):
        N, C, T, V = x.shape
        qkv = self.qkv(x).reshape(N, 3, self.num_heads, self.dim // self.num_heads, T, V).permute(1, 0, 4, 2, 5, 3)
        q, k, v = qkv[0], qkv[1], qkv[2]

        # Deriving hyperedge representation
        e = x@self.H/torch.sum(self.H, dim=0, keepdim=True)
        e = self.e_proj(e)
        e_aug = e@self.H.transpose(0, 1)
        e_aug = e_aug.reshape(N, self.num_heads, self.dim // self.num_heads, T, V).permute(0, 3, 1, 4, 2)
        
        pos_emb = self.rpe_table[self.hops]
        r = pos_emb.view(V, V, self.num_heads, self.dim // self.num_heads)
        
        a = q @ k.transpose(-2, -1)
        b = torch.einsum("bthnc, nmhc->bthnm", q, r)
        c = torch.einsum("bthnc, bthmc->bthnm", q, e_aug)
        d = torch.einsum("hc, bthmc->bthm", self.u, e_aug).unsqueeze(-2)
        
        attn = a + b + c + d
        attn = attn * self.scale
        attn = attn.softmax(dim=-1)
        x = (attn + self.relational_bias) @ v
        x = x.transpose(3, 4).reshape(N, T, -1, V).transpose(1, 2)
        x = self.proj(x)

        return x
\end{lstlisting}
\end{listing*}

\section{More experiment details}
We show in \cref{tab:a} the default hyperparameters for training our  Hyperformer on NTU RGB+D, NTU RGB+D 120 and Northwestern-UCLA datasets. We train the same 10-layer model with a total number of 216 channel dimensions for all the experiments in our paper. In addition, we provide the results of a wider model with 10 layers and 324 channels in \cref{sec:b}.

\begin{table}[ht]
  \caption{Default hyperparameters for our GRA Transformer on NTU RGB+D, NTU RGB+D 120 and Northwestern-UCLA.}
    \centering
    \begin{tabular}{lcc}
       \hline
        Config. & NTU RGB+D and NTU RGB+D 120 & Northwestern-UCLA \\
        \hline
        \hline
      random choose & False & True \\
      random rotation & True & False \\
      window size & 64 & 52\\
      weight decay & 4e-4 & 0 \\
      base lr & 2.5e-2 & 2.5e-2 \\
      lr decay rate & 0.1 & 0.1 \\
      lr decay epoch & 110, 120 & 110 120 \\
      warm up epoch & 5 & 5 \\
      batch size & 64 & 16 \\
      num. epochs & 140 & 150\\
      optimizer & Nesterov Accelerated Gradient & Nesterov Accelerated Gradient \\
       \hline
    \end{tabular}
    \label{tab:a}
\end{table}



\section{More experiment results}
\label{sec:b}
\subsection{Accuracy using single modalities}
The performance of our Hyperformer trained on joint modality only is also remarkable. We provide the experiment results for each modality on different benchmarks in detail, see
Tab. 2.

\begin{table}[h]
\centering
  \begin{tabular}{@{}l|ccccccc@{}}
    \toprule
       \multirow{2}*{Modality} & Model Size & \multicolumn{2}{c}{NTU-RGB+D 120} & \multicolumn{2}{c}{NTU-RGB+D} & \multirow{2}*{Northwestern-UCLA($\%$)}\\
       & &X-Sub(\%) & X-Set(\%)    & X-Sub(\%) & X-View(\%)   \\
    \midrule
   
    Joint & \multirow{4}*{2.6M} & 86.6  & 88.0  & 90.7  & 95.1 & 94.4\\
    Bone & & 88.0 &  89.0 & 91.2 &  95.2 & 94.6\\
    Motion &&  81.8 &  83.9 & 88.5   & 93.3 & 93.3 \\
    Bone Motion& & 82.2 &  83.5 &  88.5 & 92.6 & 92.7 \\
    \midrule
    Ensembled & &89.9 & 91.3 & 92.9 & 96.5 & 96.7 \\
    \bottomrule
  \end{tabular}
  \caption{Classification accuracy of our Hyperformer using different modalities on the NTU RGB+D, NTU RGB+D 120 and Northwestern-UCLA dataset.}
\end{table}